# Is $F_1$ Score Suboptimal for Cybersecurity Models? Introducing $C_{score}$, a Cost-Aware Alternative for Model Assessment

Manish Marwah
*OpenText*
USA

Asad Narayanan
*OpenText*
Canada

Stephan Jou
*OpenText*
Canada

Martin Arlitt
*OpenText*
Canada

Maria Pospelova
*OpenText*
Canada

*Abstract*—The cost of errors related to machine learning classifiers, namely, false positives and false negatives, are not equal and are application dependent. For example, in cybersecurity applications, the cost of not detecting an attack is very different from marking a benign activity as an attack. Various design choices during machine learning model building, such as hyperparameter tuning and model selection, allow a data scientist to trade-off between these two errors. However, most of the commonly used metrics to evaluate model quality, such as $F_1$ score, which is defined in terms of model precision and recall, treat both these errors equally, making it difficult for users to optimize for the actual cost of these errors. In this paper, we propose a new cost-aware metric, $C_{score}$, based on precision and recall that can replace $F_1$ score for model evaluation and selection. It includes a cost ratio that takes into account the differing costs of handling false positives and false negatives. We derive and characterize the new cost metric, and compare it to $F_1$ score. Further, we use this metric for model thresholding for five cybersecurity related datasets for multiple cost ratios. The results show an average cost savings of 49%.

## I. INTRODUCTION

Applications of machine learning in cybersecurity are widespread and rapidly growing, with models being deployed to prevent, detect, and respond to threats such as malware, intrusion, fraud, and phishing. The main metric for assessing the performance of classification models is $F_1$ score [1], also known as $F_1$ measure, which is the harmonic mean of precision and recall.

While $F_1$ score is used for assessing models, it is not directly used as a loss function since it is not differentiable (or convex). A simpler and commonly used approach is a two-stage optimization process. First, a model is trained using a conventional loss function such as cross-entropy, and then an optimal threshold is selected based on $F_1$ score [2].

$F_1$ score works particularly well in highly imbalanced settings prevalent in cybersecurity. However, it treats both kinds of errors a machine learning classifier can make – false positives (FPs) and false negatives (FNs) – equally. Usually, the cost of these errors is unequal and depends on the application context. For example, in cybersecurity applications, while both these errors can have severe negative consequences, one error might be preferred over the other. Specifically, false positives lead to alarm fatigue, a phenomenon where a high frequency of false alarms causes operators to ignore or dismiss all alarms. This problem is often exacerbated by base rate fallacy, where people underestimate the potential volume of false positives due to a high true positive rate while ignoring a low base rate [1]. False negatives, on the other hand, imply that a vulnerability or attack has gone undetected. While ideally one would want to minimize both these errors, in practice there is a trade-off between the number of FPs and FNs. An organization, based on its goals and requirements may assign differing costs to these errors. For example, for a ransomware detection model the damage caused by a FN may be several orders of magnitude greater than the cost of a security analyst handling a FP; while for other models, e.g., port scanning detection, the costs may be similar or even higher for handling a FP. By using $F_1$ score such cost considerations are usually ignored[2]. So a natural question to ask is: given the cost difference (or ratio) between the consequences of a FP and a FN for an particular use case, how can an organization incorporate that information while building machine learning models for that application?

There is considerable prior work in cost-sensitive learning [8]–[12], [16]. These aim to modify the model learning process, e.g., by altering the loss function to incorporate cost, adding weights to the training samples, or readjusting class frequencies in the train set such that the trained model intrinsically produces results that are cost sensitive. In this paper, we do not change the underlying learning process and instead propose a new cost-aware metric as a replacement of $F_1$ score that can be used for model thresholding, comparison

Corresponding author: Manish Marwah, mmarwah@opentext.com

---

[1]Even when the true positive rate (TPR), that is, $P(A|V)$ is high, the probability that an alarm corresponds to a real threat or vulnerability, that is, $P(V|A)$, is usually very low. This follows directly from Bayes rule: $P(V|A) \propto P(A|V) \cdot P(V)$ and the fact that the base rate, $P(V)$, is usually very low.

[2]A weighted version of $F_1$ score exists, however, it is usually not used, since it is not obvious how precision and recall should be weighed to incorporate the differing costs of FNs and FPs.

and selection. It is defined in terms of recall, precision, and a cost ratio, and can be used, for example, in determining the minimum cost point on a precision-recall curve. We applied the new metric, called cost score, $C_{score}$, to several cybersecurity related datasets and discovered significant cost differences between using $F_1$ score and $C_{score}$.

While cost score applies to any classification problem, it is especially relevant in cybersecurity since the mismatch in the costs of misclassification can be significant. The main purpose of cost score is to make it easier for practitioners to incorporate cost during model thresholding and selection. It is an easy replacement for $F_1$ score since $C_{score}$ is also defined in terms of precision and recall (and an additional cost ratio).

The key contributions of the paper are:

- Introduction of a new cost-based metric, $C_{score}$, defined as $(\frac{1}{\text{Precision}}-1-r_c)\cdot\text{Recall}+r_c$, where $r_c$ is the cost ratio, which incorporates the differing costs of misclassification and can be used as a cost-aware alternative to $F_1$ score.
- Characterization and derivation of the new metric, and its comparison with $F_1$ score.
- Application of $C_{score}$ to five cybersecurity related datasets, four of which are publicly available and one is private, for multiple values of cost ratio. The results show a cost saving of up to 86% with an average saving of 49% over using $F_1$ score in situations where costs are unequal.

## II. RELATED WORK

### A. Drawbacks of $F_1$ score and alternatives

While $F_1$ score is preferred to any one of accuracy, precision, or recall, especially for an imbalanced dataset, its primary drawback in our context is that all misclassifications are considered equal [3]. The other drawbacks [4], [5] include 1) lack of symmetry with respect to class labels, e.g., changing the positive class in a binary classifier produces a different result; and, 2) no dependence on true negatives. A more robust though not as popular alternative addressing some of these problems while still working well for imbalanced datasets is the Matthew Correlation Coefficient (MCC) [7], which in many cases is preferred to $F_1$ score [6]. It is symmetric and produces a high score only if all four confusion matrix entries (see Table II) show good results [6]. However, it treats FPs and FNs equally. Unlike $F_1$ score and MCC, our proposed metric is not symmetric with respect to FNs and FPs, taking their distinct impacts into consideration through a cost ratio. Further, like MCC but unlike $F_1$ score, our metric is symmetric in the treatment of true positives and true negatives. Our metric is not normalized like MCC and $F_1$ score, and varies between 0 (best) and $\infty$ (worst). This does not impact model thresholding, or comparison, however, the actual value of the cost metric in itself is not very meaningful, but can be converted to the corresponding recall and precision values. Since neither MCC nor $F_1$ score considers differing costs of errors, and the latter is more widely used, we compare our proposed metric with $F_1$ score in the rest of the paper.

### B. Cost sensitive learning and assessment

Since in real-world applications cost differences between types of errors can be large, cost-sensitive machine learning has been an active area of research since the past few decades [8]–[11], especially in areas such as security [14], [15] and medicine [13]. For example, Lee et al. [14] proposed cost models for intrusion detection; Liu et al. [15] incorporate cost considerations both in feature selection and classification for software defect prediction. Some of this and similar work could be used to estimate cost ratios for our proposed cost metric.

At a high-level, cost-sensitive machine learning [16] can be categorized into two different approaches: 1) where the machine learning methods are modified to incorporate the unequal costs of errors; and, 2) where existing machine learning models – trained with cost oblivious methods – are converted into cost-sensitive ones using a wrapper [10], [11]. In this paper, we focus on the second approach, which is also referred to as cost-sensitive meta learning. While there are various methods to implement this approach, we will focus on thresholding or threshold adjusting [11], where the decision threshold of a probabilistic model is selected based on a cost function. Sheng et al. [11] showed that thresholding outperforms several other cost sensitive meta learning methods such as MetaCost [10].

In the most general case, the cost function for thresholding can be constructed from the entries of a confusion matrix with a weight attached to each of them, that is, FPs, FNs, TPs and TNs [12]. Our proposed cost metric uses a similar formulation, however, it is expressed in terms of precision and recall, metrics that data scientists already know well and understand. We are not aware of any existing cost metric defined in terms of precision, recall, and a cost ratio. Unlike $F_1$ score or MCC, the proposed metric is directly proportional to the total cost of misclassification. We believe it can serve as a cost-aware replacement for $F_1$ score or MCC.

## III. PROPOSED METRIC: COST SCORE

While the proposed metric is applicable to any machine learning classification model, including multiclass and multi-label settings, for simplicity we will assume a binary classification task in the following discussion. The notation used is summarized in Table I.

Starting with the cost of misclassifications, we derive expressions for cost score that can replace $F_1$ score. In particular, we derive two equivalent expressions — one in terms of TPR (recall) and FPR; the other in terms of precision and recall. They both include an error cost ratio ($r_c$), which is a ratio between the cost of a FN to that of a FP. The first one is dependent on the base rate ($P(V)$), while the second, similar to F1-score, does not directly depend on it.

The basic evaluation metrics for a binary classifier can be defined from a confusion matrix, shown in Table II. One can also look at a confusion matrix from a probabilistic perspective, where the four possible outcomes define a probabilistic space, with each outcome a joint probability, as shown in Table

III. The total probability along a row or a column are the corresponding marginal probabilities.

TABLE I: Notation

| Symbol | Description |
|---|---|
| $\neg$ | logical not |
| $V$ | vulnerability or threat, or in general positive class |
| $A$ | positive classification by a detector, which may result in an alarm |
| $TP$ | true positive |
| $TN$ | true negative |
| $FP$ | false positive |
| $FN$ | false negative |
| $N$ | total number of data points |
| $N_{FP}$ | number of false positives |
| $N_{FN}$ | number of false negatives |
| $p$ | total number of positives |
| $\hat{p}$ | total number of predicted positives |
| $n$ | total number of negatives |
| $\hat{n}$ | total number of predicted negatives |
| $C_{FP}$ | cost of a false positive |
| $C_{FN}$ | cost of a false negative |
| $C$ | total cost of misclassification |
| $r_c$ | Error cost ratio, defined as $\frac{C_{FN}}{C_{FP}}$ |
| $R$ | Recall |
| $Prec$ | Precision |

TABLE II: Confusion Matrix

| | | Ground Truth | | |
|---|---|---|---|---|
| | | $V$ (or $T$) | $\neg V$ (or $F$) | |
| Prediction | $A$ (or $T$) | TP | FP | $\hat{p}$ |
| | $\neg A$ (or $F$) | FN | TN | $\hat{n}$ |
| | | p | n | |

TABLE III: Confusion Matrix – probabilistic view

| | | Ground Truth | | |
|---|---|---|---|---|
| | | $V$ | $\neg V$ | |
| Prediction | $A$ | $P(A,V)$ | $P(A,\neg V)$ | $P(A)$ |
| | $\neg A$ | $P(\neg A,V)$ | $P(\neg A,\neg V)$ | $P(\neg A)$ |
| | | $P(V)$ | $P(\neg V)$ | |

*Conditional Probabilistic Definitions of Classifier Metrics*

**False positive rate** ($\frac{FP}{n}$): $P(A|\neg V)$

**False negative rate** ($\frac{FN}{p}$): $P(\neg A|V)$

**True positive rate (recall)** ($\frac{TP}{p}$): $P(A|V)$

**True negative rate** ($\frac{TN}{n}$): $P(\neg A|\neg V)$

**Precision** ($\frac{TP}{\hat{p}}$): $P(V|A)$

**False discovery rate (or 1 - precision)** ($\frac{FP}{\hat{p}}$): $P(\neg V|A)$

*A. Cost function*

The cost incurred as a result of misclassification is composed of the cost of false positives and that of false negatives. $P(A, \neg V)$ and $P(\neg A, V)$ represent the probability of false positives and false negatives, respectively. Thus, their number can be expressed as:

$$N_{FP} = N \cdot P(A, \neg V)$$
$$N_{FN} = N \cdot P(\neg A, V)$$

Multiplying by the corresponding costs gives us the total cost of errors:

$$C = C_{FP} \cdot N_{FP} + C_{FN} \cdot N_{FN}$$
$$= C_{FP} \cdot N \cdot P(A, \neg V) + C_{FN} \cdot N \cdot P(\neg A, V)$$

Factoring out the common terms and introducing $r_c$ gives:

$$C = C_{FP} \cdot N(P(A, \neg V) + r_c \cdot P(\neg A, V)) \quad (1)$$
$$= K \cdot [P(A, \neg V) + r_c \cdot P(\neg A, V)] \quad (2)$$

*B. Cost score in terms of TPR and FPR*

Here we model the cost function in terms of TPR (recall) and FPR. Data scientists frequently evaluate a model in terms of TPR, which corresponds to the fraction of positive cases detected and FPR, which is the fraction of the negatives that were misclassified as positives. In fact, an ROC curve (a plot between TPR and FPR) is widely used for thresholding a probabilistic classifier.

Using the product rule, we rewrite the probability distribution for false positives in terms of FPR and $P(V)$.

$$P(A, \neg V) = P(A|\neg V) \cdot P(\neg V) \quad (3)$$
$$= FPR \cdot (1 - P(V)) \quad (4)$$

Similarly, we rewrite the joint distribution for false negatives in terms of TPR and $P(V)$.

$$P(\neg A, V) = P(\neg A|V) \cdot P(V) \quad (5)$$
$$= (1 - P(A|V)) \cdot P(V) \quad (6)$$
$$= (1 - TPR) \cdot P(V) \quad (7)$$

Substituting 4 and 7 in the cost expression, 2, and rearranging, we get:

$$C = K \cdot [FPR + P(V)(r_c - r_c \cdot TPR - FPR)]$$

To minimize the cost, we can ignore $K$, and thus the cost score becomes:

$$C_{score} = FPR + P(V) \cdot (r_c - r_c \cdot TPR - FPR) \quad (8)$$

## C. Cost score in terms of precision and recall

While FPR is a useful metric as it captures the number of false positives, it can be tricky to understand, especially when the base rate, $P(V)$, is low, which is usually the case in cybersecurity problems. For problems such as intrusion or threat detection, FPs add overhead to the workflow of a security analyst. For phishing website detection, a FP may result in a website being blocked in error for an end user. In either case, setting a target FPR requires knowledge of the base rate and would change as the base rate changes. In other words, even a seemingly low FPR may not be good enough, given a low base rate. Further, variance in base rate would affect overhead of a security analyst in case of intrusion detection or the fraction of erroneously blocked websites for a user in case of phishing detection *even if* the FPR stays constant. Precision on the other hand directly captures the operator overhead or fraction of erroneously blocked websites independent of the base rate.

A main attraction of $F_1$ score is its use of precision instead of FPR. When the costs of FP and FN are similar, $F_1$ score is an effective evaluation metric, however, with unequal costs of misclassification, we can usually find a better solution by incorporating this cost differential in the metric. Below, we derive an expression for $C_{score}$ in terms of precision and recall, similar to $F_1$ score, but that includes a cost ratio.

We can rewrite the probability of a false positive in terms of precision ($Prec$) and marginal probability of alarm.

$$P(A, \neg V) = P(\neg V | A) \cdot P(A) \quad (9)$$
$$= (1 - Prec) \cdot P(A) \quad (10)$$

P(A) can be expressed in terms of $P(V)$, $Prec$ and $R$ (recall) using Bayes rule:

$$P(V|A) \cdot P(A) = P(A|V) \cdot P(V)$$
$$P(A) = \frac{P(A|V)}{P(V|A)} \cdot P(V)$$
$$= \frac{R}{Prec} \cdot P(V)$$

Substituting this value of $P(A)$ in Equation 10, we get:

$$P(A, \neg V) = \frac{1 - Prec}{Prec} \cdot R \cdot P(V) \quad (11)$$

As in the previous section (Equation 7), the probability of a false negative can be written as:

$$P(\neg A, V) = (1 - R) \cdot P(V) \quad (12)$$

Therefore, substituting the values of probabilities of a false positive and a false negative from Equations 11 and 12, respectively, into the cost expression (Equation 1), we get

$$C_{score} = N \cdot C_{FP} \cdot P(V) \cdot [\frac{1 - Prec}{Prec} \cdot R + r_c(1 - R)]$$

Since $N$, $C_{FP}$ and $P(V)$ are constant for a given dataset, we can rewrite the cost expression as:

$$C_{score} = (\frac{1}{Prec} - 1) \cdot R + r_c(1 - R) \quad (13)$$

This expression defines the cost in terms of precision, recall and cost ratio and can be used instead of $F_1$ score for any tasks that require model comparison such as model thresholding, hyperparameter tuning, model selection and feature selection.

$C_{score}$ goes to zero for $Prec = 1$ and $R = 1$, as expected. As $Prec \to 0$ and $R \to 0, C_{score} \to \infty$.

We have derived two equivalent cost expressions – one involving TPR and FPR (quantities used in an ROC curve) and the second involving precision and recall (quantities used in computing $F_1$ score). Similarly, it may be possible to derive additional equivalent cost expressions in terms of other commonly used metrics. In the remainder of the paper, we will only consider the cost expression $C_{score}$ defined in terms of precision and recall (similar to $F_1$ score). This definition of $C_{score}$ is not directly dependent on the base rate ($P(V)$), unlike the one in the previous section.

## D. $C_{score}$ Isocost Contours

To better understand the cost score metric, we will examine its dependence on precision and recall, and compare it with $F_1$ score. Figure 1 shows a precision-recall (PR) plot with $F_1$ score isocurves or contours. Each curve corresponds to a constant value of $F_1$ score as specified next to the curve. If recall and precision are identical, $F_1$ score computes to the same value. However, if there is a wide gap between them, $F_1$ tends to be closer to the lower value, as can be seen in the top-left and bottom-right regions of the plot. As expected, the highest (best) value contours are towards the top-right corner of the plot (that is, towards perfect recall and precision). Further, the slope of the curves is always negative (as shown in Appendix A), implying there is always a trade-off between recall and precision.

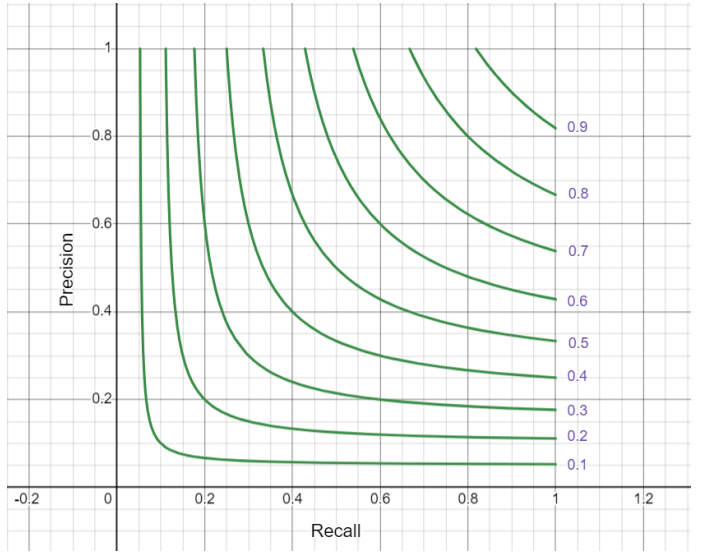

Fig. 1: Precision-Recall plot of F1-score iso-curves.

We can similarly obtain isocost curves (or contour lines) for cost score assuming a particular cost ratio, $r_C$. The cost score (Equation 13) can written as:

$$C_{score} = (\frac{1}{Prec} - 1 - r_c) \cdot R + r_c \quad (14)$$

and plotted for constant values of $C_{score}$ on a PR plot. Figure 2 shows the isocost curves for three cost ratios: $r_c = 1$, that is, FN and FP cost the same; $r_c = 10$, that is, FN are ten times as expensive as FP; and $r_c = 0.1$, that is, FN are one-tenth as expensive as FP.

There are three distinct regions in the plot, based on the slope of the curves. From the above equation, we can compute the slope (see Appendix, A for details).

$$\frac{\partial Prec}{\partial R} = \frac{C_{score} - r_c}{(C_{score} + R(r_c + 1) - r_c)^2}$$

Depending on the value of $C_{score}$, the slope can be positive, negative or zero as shown below.

$$\frac{\partial Prec}{\partial R} = \begin{cases} > 0 & \text{if } C_{score} > r_c \\ < 0 & \text{if } C_{score} < r_c \\ = 0 & \text{if } C_{score} = r_c \end{cases}$$

For lower (better) values of $C_{score}$, when $C_{score} < r_c$, the slope is negative and the isocost curves are similar to the isocurves for $F_1$ score. The horizontal line corresponds to $C_{score} = r_c$, and the curves below it have a positive slope with $C_{score} > r_c$. The isocurves closest to the top-right corner have the lowest costs.

While the isocost contours are plotted assuming $Prec$ and $R$ are independent, that is obviously not the case for a particular model. In fact, $Prec, P(V|A)$, and $R, P(A|V)$, are related by Bayes rule: $Prec = \frac{P(V)}{P(A)} \cdot R$ The feasible $Prec$-$R$ pairs obtained by varying model thresholds are given by a PR curve. A hypothetical PR curve is shown as a dotted black line in Figure 2. The cost corresponding to each point on the PR curve is given by the isocost intersecting that point. The minimum cost point on the PR curve is the one that intersects the lowest cost contour. If the PR curve is convex, the minimum cost contour will touch the PR curve only at one point where their tangents have equal slope[3]. However, in practice empirically constructed PR curves are not always convex and thus the minimum cost point may not be unique. In Figure 2 point A, B and C approximately show the minimum cost point for the three cost ratios.

What do isocost contours mean in terms of the confusion matrix? $C_{score}$ remains constant along a contour, and is proportional to $FP + r_c FN$, which must remain constant as recall and precision change. In Table IV, we have parameterized the confusion matrix entries with $k$ such that as $k$ changes for a particular $r_c$, precision and recall vary, however, $C_{score}$ remains constant. This can be seen by computing $FP + r_c FN$ for the Table entries, which is $(FP' + r_c k) + r_c(FN' - k)$ and independent of $k$ and thus constant.

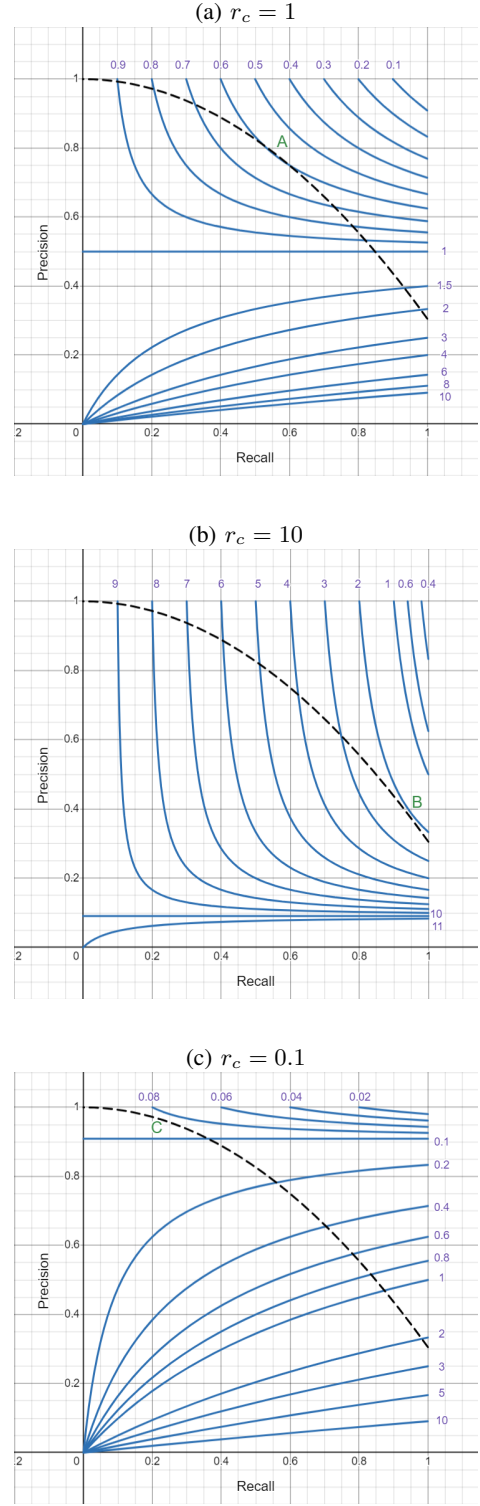

Fig. 2: Isocost contours for $C_{score}$ for three different cost ratios, $r_c$. The $C_{score}$ corresponding to each contour is listed next to it. The black dotted-line is the PR curve for a particular model.

---

[3] Under assumption of convexity, this can be proved by contradiction. Assume the minimum isocost touches a PR curve at at least two points; however, since both functions are convex, there must be another lower cost isocost touching the PR curve at at least one point. Thus, the lowest cost isocost must touch the PR curve at exactly one point.

TABLE IV: Confusion Matrix – parameterized by $k$

|  |  | Ground Truth | | |
|---|---|---|---|---|
|  |  | $V$ | $\neg V$ | |
| Prediction | $A$ | $TP' + k$ | $FP' + r_c \cdot k$ | $\hat{p} + r_c \cdot k + k$ |
|  | $\neg A$ | $FN' - k$ | $TN' - r_c \cdot k$ | $\hat{n} - r_c \cdot k - k$ |
|  |  | p | n | |

### E. How does $F_1$ score Compare with $C_{score}$ when $r_c = 1$?

While $F_1$-score varies from 0 to 1, with 1 indicating perfect performance, $C_{score}$ is proportional to the actual cost of handing model errors, with a zero-cost indicating perfect performance (that is, no FPs or FNs). $F_1$ score treats FNs and FPs uniformly, as does $C_{score}$ when $r_c = 1$. So a natural question is if $C_{score}$ differs from $F_1$ score when $r_c = 1$? To compare, we transform $F_1$-score to a cost metric:

$$F1_{cost} = \frac{1}{F_1} - 1 \quad (15)$$

When $F_1$ is 1, $F1_{cost} = 0$, and when $F_1$ is 0, $F1_{cost} \to \infty$, and thus exhibits behavior of a cost function and can be directly compared with $C_{score}$.

To compare $C_{score}$ and $F1_{cost}$, we reduce both in terms of the elements of the confusion matrix and find that:

$$C_{score} \propto FP + FN \quad (16)$$

$$F1_{cost} \propto \frac{FP + FN}{TP} \quad (17)$$

Thus, when $r_c = 1$, $C_{score}$ and $F1_{cost}$ are not identical; while $C_{score}$ is proportional to the total number of errors, $F1_{cost}$ is also inversely proportional to the number of true positives. $C_{score}$ only considers the cost of errors; it assigns zero cost to both TPs and TNs. In that sense, it treats TP and TN symmetrically unlike $F_1$ score.

### F. Multiclass and multilabel classifiers

While we derived the cost metric assuming a binary classification problem, its extension to multiclass and multilabel classification problems is straightforward. A cost ratio per class would need to be defined. For a multiclass classifier, a user would have to assign cost ratios considering each class as positive and the rest as negative. Similarly, for a multilabel classifier, a user would have to assign independent ratios for each class. This will allow a $C_{score}$ to be computed per class. To compute a single cost metric, the per-class $C_{score}$ would need to be aggregated. The simplest aggregation function is an arithmetic mean, although a class weighted mean based on class importance, or another type of aggregation, e.g., a harmonic mean, can also be performed. The "one class versus the rest" approach is similar to how $F_1$ score and other metrics are computed in a multiclass setting.

### G. Minimizing $C_{score}$ based on model threshold and other hyperparameters

In Section III-D, we described use of isocost contours to visually determine the lowest cost point on a PR curve. In practice, to find the minimum cost based on model threshold, and the corresponding precision-recall values, precision and recall can be considered functions of the threshold value ($t$) with the optimal value of $t$ determined by minimizing the cost function with respect to $t$.

$$t = \text{argmin}_t [(\frac{1}{Prec(t)} - 1) \cdot R(t) + r_c \cdot (1 - R(t))]$$

In addition to model threshold, $C_{score}$ can also be used for selecting other model hyperparameters such as the number of neighbors in $k$-NN; number of trees, maximum tree depth, etc. in tree based models; number and type of layers, activation functions, etc. in neural networks; and for model comparison and selection. Hyperparameter tuning [22] is typically performed using methods such as grid search, random search, gradient-based optimization, etc. Typically, cross-validation is used in conjunction to evaluate the quality of a particular choice of a dataset. In all these methods, the proposed $C_{score}$ can replace another cost-oblivious metric such as $F_1$ score.

## IV. EXPERIMENTAL EVALUATION

### A. Datasets

The datasets used for experiments were chosen based on their relevance to security and the varying cost of misclassification between target classes. To comprehensively analyze the impact of costs, we selected five different datasets, four publicly available datasets and one privately collected dataset. The publicly available datasets include the UNSW-NB15 intrusion detection data, KDD Cup 99 network intrusion data, credit card transaction data, and phishing URL data.

1) **UNSW-NB15 Intrusion Detection Data**: This network dataset, developed by the Intelligent Security Group at UNSW Canberra, comprises events categorized into nine distinct types of attacks including normal traffic. To suit the experimental requirements of our study, the dataset was transformed into a binary classification setting, where a subset of attack classes (Backdoor, Exploits, Reconnaissance) are consolidated into class 1, while normal traffic is represented as class 0. There are a total of 93,000 events in class 0 and 60,841 events in class 1. For our research, we utilized the CSV version of the dataset, which comes pre-partitioned into training and testing sets [18] [19].

2) **KDD Cup 99 Network Intrusion Data**: This dataset originated from packet traces captured during the 1998 DARPA Intrusion Detection System Evaluation. It encompasses 145,585 unique records categorized into 23 distinct classes, which include various types of attacks alongside normal network traffic. Each record is characterized by 41 features that are derived from the packet traces. For this research, the dataset has been adapted to focus on a binary classification task: class 0 represents normal instances, while class 1 aggregates all other attack types. Also, to explore the impact of different thresholds on the model's performance, training was conducted using only 1% of the dataset. The dataset is accessed through the datasets available in the Python sklearn package [21].
3) **Credit Card Transactions Data**: This dataset contains credit card transaction logs with 29 features that are tagged into legitimate and fraudulent transactions. There are a total of 284,315 transactions out of which 492 are fraudulent (class 1) and 56,866 are legitimate (class 0) [17]. Of the five datasets, this one has the highest skew.
4) **Phishing data**: This dataset is a collection of 60,252 webpages along with their URL and HTML sources. Out of these, 27,280 are phishing sites (class 1) whereas 32,972 are benign (class 0) [20]. We only use the URLs for building the model.
5) **Internal data**: This is a private dataset, used within an organization that represents the results of an extensive audit conducted on vulnerabilities in source code. Each vulnerability is classified as into two classes class 0 or class 1 (actual class names are masked for anonymity) by human auditors during the auditing process. The model is trained on this manually audited data and predicts if a given vulnerability belongs to class 0 or class 1. There are a total of 144,978 instances out of which 18,738 belong to class 1. Each vulnerability has 58 features which encompass a wide array of metrics that were generated during the analysis of the codebase.

TABLE V: Summary of Datasets

| Dataset | Number of instances | | Number of features |
| --- | --- | --- | --- |
| | Class 0 | Class 1 | |
| UNSW-NB15 | 93,000 | 60,841 | 42 |
| Credit card fraud | 284,315 | 492 | 29 |
| KDD cup 99 | 87,832 | 57,753 | 41 |
| Phishing data | 32,972 | 27,280 | 188 |
| Internal data | 126,240 | 18,738 | 58 |

The information about each dataset is summarized in Table V. It is important to note that not all datasets used are balanced. For instance, credit card fraud data has less than 1% of instances in class 1. Similarly, the internal data has only 15% instances in class 1.

### B. Experiment Setup

We train a classification model using a RandomForest algorithm for each dataset. The goal is not to train the best possible model for the dataset but to obtain a reasonably good model with a probabilistic output.

The steps are as follows:
1) **Model Training**: A RandomForest classifier is trained on each dataset. Although the training sets have different skews, we effectively used a balanced dataset for training so the classifier gets an equal opportunity to learn both classes.
2) **Threshold adjustment using $F_1$ score**: The validation dataset is used to identify the best threshold based on the $F_1$ score. The validation dataset was selected by sampling a proportion of the data, ensuring that the class distribution mirrored that of the training data. This approach was taken because the actual skew of classes in production deployment is unknown. However, it is important to note that for actual production systems, the validation set should be representative of the true data distribution. Specifically, for the UNSW-NB15 dataset, the validation set was sampled from events in the test data CSV file.
3) **Threshold adjustment using $C_{score}$**: The predictions from the trained model are analyzed across different cost ratios. Using the validation sets, we apply the $C_{score}$ to determine the optimal threshold for each cost ratio.
4) **Comparison**: The model's cost with thresholds chosen based on the $F_1$ score is compared against the costs with thresholds chosen using the $C_{score}$.

This setup allows us to evaluate the effectiveness of $C_{score}$ in optimizing model performance under varying cost conditions.

### C. Results

The proposed cost metric used is tailored for enhancing the performance of machine learning models in scenarios where the cost of false negatives greatly differs from the cost of false positives. This in turn helps in optimizing predictions based on cost considerations, thereby addressing a critical limitation in existing evaluation methods.

*1) $F_1$ Score for thresholding*: To illustrate the advantages of our approach, we will first use $F_1$ score to adjust a model's threshold. Figure 3 depicts the changes in $F_1$ score for different threshold values across each dataset. The histograms in each plot represent the distribution of data within the corresponding probability intervals. Each color in the histogram represents the distribution of the corresponding ground truth class represented by the color. Due to the significant skew of the credit card dataset, the density of class 1 is not visible in the histogram.

The $F_1$ score for each dataset starts from a threshold of 0, where all instances are classified as the positive class and recall is 1. It ends at a score of 0 at a threshold of 1, where all instances are tagged as negatives and recall is 0. The rate of change of the $F_1$ score in a threshold interval is proportional to the proportion of data within the interval and their ground truth values. This explains why, for some datasets, the $F_1$ score is flat or nearly flat in the middle range of thresholds.

It is clear from Figure 3 that most models do a good job of separating the two classes. The model trained on KDD cup 99 data is able to separate both the classes more distinctively and has most of the data points near probability zero and one. This makes the threshold vs $F_1$ score curve mostly flat in the middle range of the probabilities. The best $F_1$ score is achieved at a threshold of 0.33. Similarly, the phishing, credit card fraud, and intrusion detection datasets perform well on the validation datasets with well-separated bimodal distributions. The threshold with the highest $F_1$ score is marked with a vertical line in each plot. For the internal dataset, the model struggles to separate both classes as can be seen by the overlap in the probabilities of both classes. The model achieves its best $F_1$ score at a threshold of 0.688.

As the threshold moves from 0 to 1 there is a tradeoff between FPs and FNs; the maximum $F_1$ score for each model corresponds to the point where the sum of FPs and FNs are minimal while the number of TPs are the highest. Being symmetrical in FN and FP, $F_1$ score reduces their sum, disregarding any class specific costs.

*2) $C_{score}$ for thresholding at different cost ratios:* The proposed $C_{score}$ metric allows the tuning of model parameters based on a cost ratio (ratio of the cost of false negatives to the cost of false positives). This cost ratio is variable and dependent on the specific impacts these errors have on end users. For example, in scenarios where missing a true attack could lead to significant financial losses, the cost of a false negative is higher. Conversely, in resource-constrained environments, a high rate of false positives can considerably burden the evaluation process.

To illustrate the tuning differences, we applied three distinct cost ratios to each dataset: 0.1 (where a false positive is ten times more costly than a false negative), 1 (equal cost for both false positives and false negatives), and 10 (where a false negative is ten times more costly than a false positive). These cost ratios are used solely to demonstrate the model's behavior when tuned with $C_{score}$ and may not correspond to practical applications of the data.

Figure 4 displays the optimal thresholds derived from $C_{score}$ for each dataset across these cost ratios. The histograms in each plot show the distribution of the ground truth classes within the probability intervals. The $C_{score}$ reflects the classification cost, resulting in a curve shape inverse of that of the $F_1$ score, with the optimal threshold at the minimum $C_{score}$. Similar to the $F_1$ score plot, the flat portions of the $C_{score}$ curve correspond to probability intervals with fewer data points. At a threshold of 0, the $C_{score}$ is constant regardless of the cost ratio, as the recall is 1 and $C_{score}$ is $\frac{1}{Prec} - 1$.

Table VI summarizes the experimental results, comparing the performance improvements of $C_{score}$ at different thresholds with those of the $F_1$ score. Cost score ($C_{score}$) is computed for thresholds based on maximizing $F_1$ score and minimizing $C_{score}$ for each of the cost-ratios. There is only one threshold for each dataset based on best $F_1$ score but the threshold based on $C_{score}$ varies for each cost ratio. Although the actual cost is a multiple of the $C_{score}$, the percentage improvement over the $F_1$ score reflects the reduction in actual cost.

For the UNSW-NB15 dataset, the optimal threshold is 0.89 for a cost ratio of 0.1, minimizing false positives at the expense of some true positives becoming false negatives (Figure 4a).

At a cost ratio of 1, the threshold decreases to 0.65, balancing false positives and false negatives, aligning closely with the best $F_1$ score threshold. At a cost ratio of 10, the threshold further decreases to 0.42, significantly reducing false negatives despite an increase in false positives.

In the phishing dataset, the probability distribution of both classes is similar (Figure 4b). At a cost ratio of 1, the threshold is 0.54, identical to the best $F_1$ score threshold. For a cost ratio of 0.1, the threshold increases to 0.73 to reduce false positives. Conversely, at a cost ratio of 10, the threshold decreases to 0.17 to significantly reduce false negatives. The spike in $C_{score}$ for a cost ratio of 10 is proportional to the number of true class 1 instances within the probability interval.

For the credit card fraud and KDD Cup 99 datasets, the $C_{score}$ curve remains mostly flat. In the case of credit card fraud data, we applied a logarithmic transformation (Figure 4c) to highlight the differences in $C_{score}$ due to the significant class imbalance. For the KDD Cup 99 dataset, the trained model achieves good class separation (Figure 4d), resulting in a relatively flat $C_{score}$ curve in the middle region, with a spike towards a probability interval of 1 as the cost ratio increases.

In the internal dataset, as we saw earlier, there is substantial overlap between the probability intervals of both classes, increasing the significance of false positives and false negatives (Figure 4e). At a cost ratio of 0.1, the threshold is set at 0.95, nearly eliminating false positives. At a cost ratio of 1, the threshold is 0.92, which is only slightly different from the threshold for a cost ratio of 0.1 and results in almost similar rates of false positives. This slight decrease can be attributed to significant class imbalances, where further threshold reduction could significantly increase false positives due to the higher count of instances in class 0. As the cost ratio increases to 10, the threshold decreases to 0.42, considerably reducing false negatives (as indicated by the reduced proportion of class 1 instances to the left of the threshold). The spike in $C_{score}$ at this cost ratio corresponds to the interval with a significant count of class 1 instances.

Table VI compares the cost improvements achieved by $C_{score}$ at different cost ratios to the costs at optimal thresholds based on the $F_1$ score. Performance improvements with respect to $C_{score}$ range from 10% to 85% in most scenarios for cost ratios of 0.1 and 10, with an average cost improvement of 49%. At a cost ratio of 1, the improvement is minimal, except in datasets with significant class imbalances, indicating the similarity between $F_1$ score and $C_{score}$ at this ratio. For the internal dataset with a cost ratio of 10, the cost improvement at the optimal $C_{score}$ threshold compared to the $F_1$ score is 86%. Additionally, there is over 50% improvement in cost at a cost ratio of 0.1 for the UNSW-NB15, credit card fraud, and KDD Cup 99 datasets, underscoring the substantial benefits of tuning models using the $C_{score}$ metric. These findings demonstrate how $C_{score}$ effectively adjusts the threshold to balance false positives and false negatives based on the specified cost ratio.

*3) **Precision-Recall trade-off using** $C_{score}$:* Cost score's ability to balance false negatives and false positives based on varying cost ratios is further demonstrated by the changes in precision and recall (Figure 5 and Table VI). The results indicate that as the cost ratio shifts from 1 to 0.1, precision increases while recall decreases. For example, in the case of UNSW-NB15 data, precision reaches 0.992 at a cost ratio of 0.1 which highlights a reduction in false positives. Conversely, when the cost ratio increases to 10, there is a significant improvement in recall. For example, in the case of UNSW-NB15 data, the recall is improved 10% compared to that at best $F_1$ score at a cost ratio of 10, indicating a reduction in false negatives.

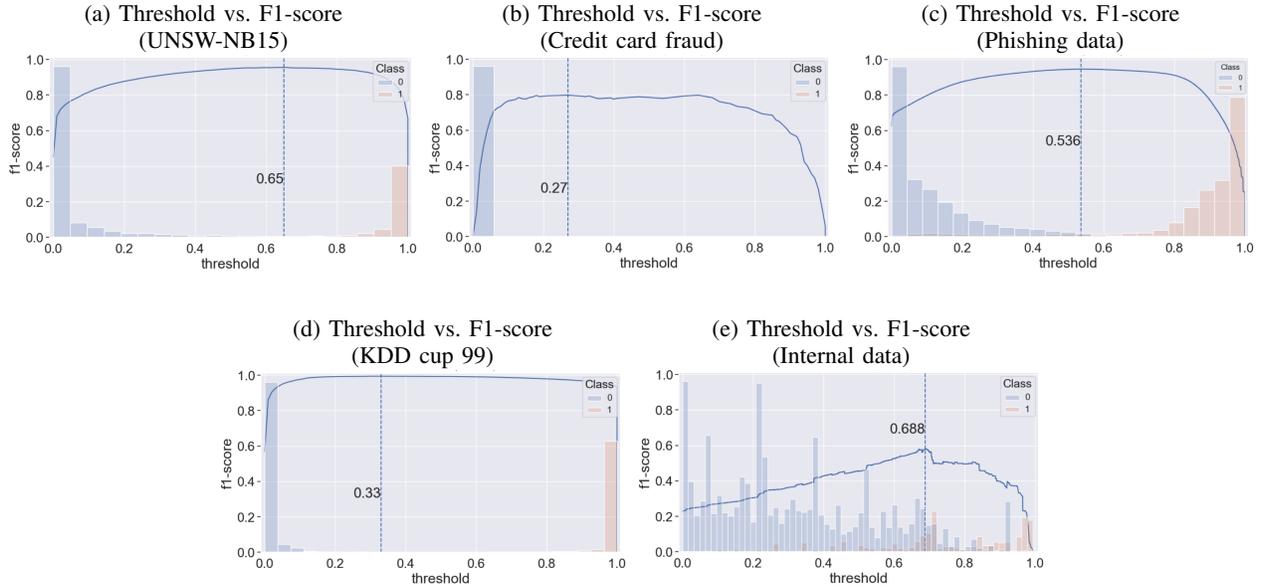

Fig. 3: Threshold for best $F_1$ score for each dataset.

TABLE VI: Misclassification costs based on using $F_1$ score and $C_{score}$ for thresholding for three different cost ratios for the five datasets

| Dataset | Cost ratio | $F_1$ score based threshold | | | | $C_{score}$ based threshold | | | | Percentage Improvement in Cost |
|---|---|---|---|---|---|---|---|---|---|---|
| | | Optimal threshold | Precision | Recall | $C_{score}$ | Optimal threshold | Precision | Recall | $C_{score}$ | |
| UNSW-NB15 | 0.1 | 0.65 | 0.949 | 0.961 | 0.056 | 0.890 | 0.992 | 0.868 | 0.020 | 64.1% |
| | 1 | | | | 0.091 | 0.650 | 0.949 | 0.961 | 0.091 | 0.0% |
| | 10 | | | | 0.441 | 0.420 | 0.885 | 0.993 | 0.203 | 53.2% |
| Credit card fraud | 0.1 | 0.27 | 0.815 | 0.781 | 0.199 | 0.900 | 0.976 | 0.417 | 0.069 | 65.3% |
| | 1 | | | | 0.396 | 0.640 | 0.931 | 0.698 | 0.354 | 10.6% |
| | 10 | | | | 2.365 | 0.130 | 0.757 | 0.812 | 2.135 | 9.7% |
| KDD cup 99 | 0.1 | 0.33 | 0.995 | 0.994 | 0.006 | 0.540 | 0.999 | 0.986 | 0.002 | 66.7% |
| | 1 | | | | 0.011 | 0.330 | 0.995 | 0.994 | 0.011 | 0.0% |
| | 10 | | | | 0.065 | 0.170 | 0.982 | 0.998 | 0.034 | 47.7% |
| Phishing data | 0.1 | 0.54 | 0.980 | 0.915 | 0.027 | 0.730 | 0.997 | 0.873 | 0.015 | 44.4% |
| | 1 | | | | 0.104 | 0.540 | 0.980 | 0.915 | 0.104 | 0.0% |
| | 10 | | | | 0.876 | 0.170 | 0.764 | 0.970 | 0.595 | 32.1% |
| Internal data | 0.1 | 0.69 | 0.532 | 0.637 | 0.597 | 0.948 | 0.971 | 0.230 | 0.084 | 85.9% |
| | 1 | | | | 0.923 | 0.923 | 0.942 | 0.252 | 0.764 | 17.2% |
| | 10 | | | | 4.186 | 0.424 | 0.292 | 0.886 | 3.289 | 21.4% |

When comparing the results of the new cost score at a cost ratio of 1 with the $F_1$ score, the outcomes are similar in most cases. This similarity is evident from the Precision-Recall curves (Figure 5), where the precision and recall for the optimal $F_1$ score overlap with those of the new cost score at a cost ratio of 1. This underscores the fact that the $F_1$ score consistently assigns equal weights to both false negatives and false positives, irrespective of their real-world cost impacts. The notable difference in precision and recall between the $F_1$ score and the new cost score is observed in the credit card fraud data (Figure 5b) and internal data (Figure 5e), which can be attributed to the significant class imbalance in this dataset and also to the fact that $F_1$ score is also proportional to true positives and thereby strives for a better recall compared to $C_{score}$ at cost ratio 1.

These results demonstrate that the proposed $C_{score}$ metric offers substantial performance enhancements over the $F_1$ score across a range of cost ratios. Unlike the $F_1$ score, which typically assigns equal penalties to false positives and false negatives, the $C_{score}$ metric accommodates variations in the costs associated with these errors. At a cost ratio of 1, the $C_{score}$ achieves performance comparable to the $F_1$ score, illustrating its versatility. The flexibility of the $C_{score}$ to tune models based on cost ratios, particularly demonstrated by the results on internal data, has proven it to be a valuable metric for fine-tuning our models to meet the varying cost demands of end users. These findings suggest that the $C_{score}$ can effectively replace the $F_1$ score for tasks such as model thresholding and selection, particularly in scenarios where the costs of false positives and false negatives differ.

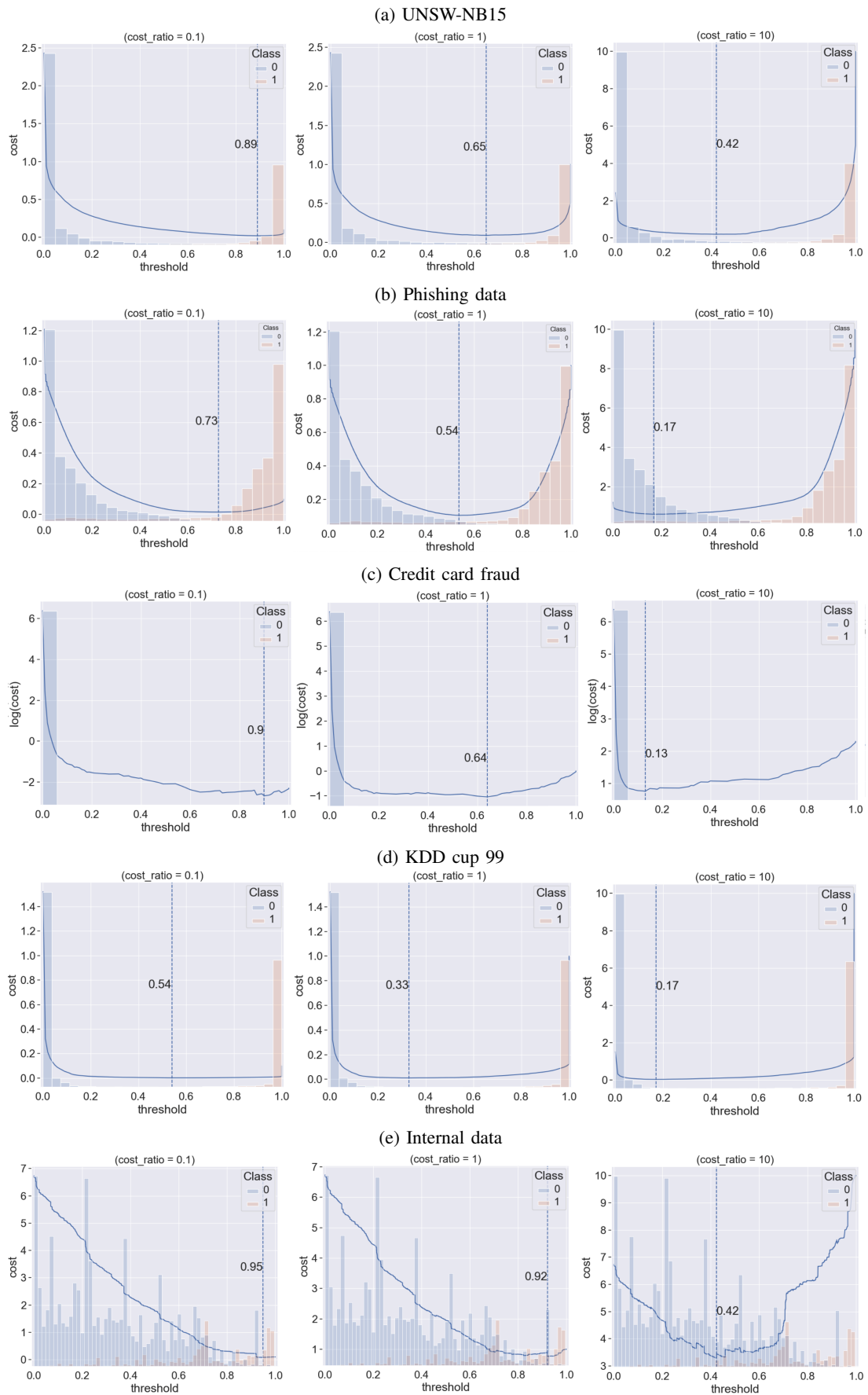

Fig. 4: Variation of thresholds with different cost ratios for each dataset.

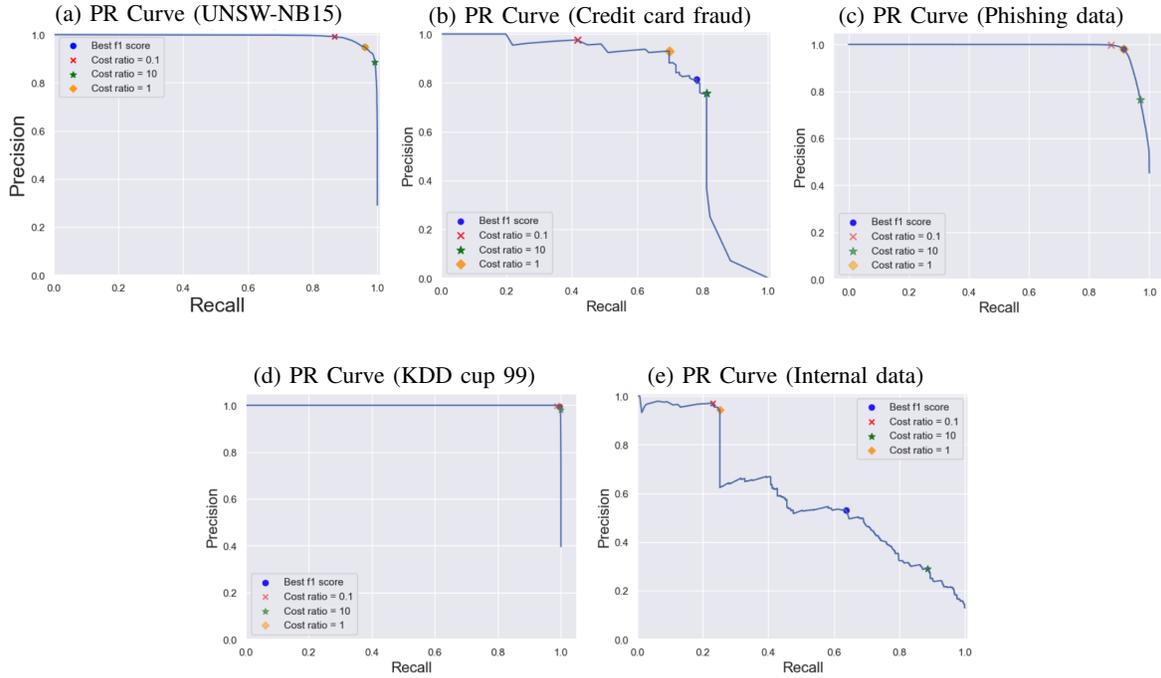

Fig. 5: Precision-Recall curve for different cost ratios

## V. CONCLUSIONS

How organizations handle errors from machine learning models is highly dependent on context and application. In the cybersecurity domain, the cost of a security analyst's time and effort spent in reviewing and investigating a false positive varies considerably from the cost of a model's failure to detect a real security incident (a false negative). However, widely used metrics like $F_1$ score assign them equal costs. In this paper, we derived a new cost-aware metric, $C_{score}$ defined in terms of precision, recall, and a cost ratio, which can be used for model evaluation and serve as a replacement for $F_1$ score. In particular, it can be used for thresholding probabilistic classifiers to achieve minimum cost. To demonstrate the effectiveness of $C_{score}$ in cybersecurity applications, we applied it to threshold models built on five different datasets assuming multiple cost ratios. The results showed substantial savings in cost through the use of $C_{score}$ over $F_1$ score. At cost ratio 1, the results are similar, however, as the cost ratio is increased or decreased, the gap in costs between using $C_{score}$ and $F_1$ score increases. All datasets show consistent improvements in cost. Through this work, we hope to raise awareness among machine learning practitioners building cybersecurity applications regarding the use of cost-aware metrics such as $C_{score}$ instead of cost-oblivious ones like $F_1$ score.

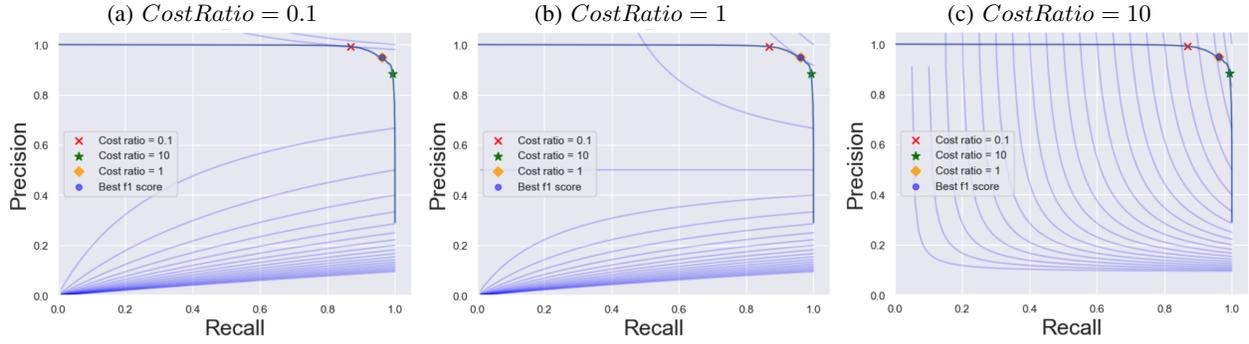

Fig. 6: Contour plots with Precision and Recalls

APPENDIX

A. Slopes of the $F_1$ score and $C_{score}$ isocurves

$F_1$ score is defined as:

$$F_1 = \frac{2 \cdot Prec \cdot R}{Prec \cdot R}$$

Rearranging:

$$Prec = \frac{F_1 \cdot R}{2R - F_1}$$

Slope of $F_1$ isocurves can be calculated as:

$$\frac{\partial Prec}{\partial R} = \frac{F_1}{2R - F_1} - \frac{2RF_1}{(2R - F_1)^2}$$
$$= \frac{-F_1^2}{(2R - F_1)^2}$$

Thus, slope of $F_1$ isocurves is always negative.

$C_{score}$ is defined as:

$$C_{score} = (\frac{1}{Prec} - r_c - 1) \cdot R + r_c$$

It can be rearranged as:

$$Prec = \frac{R}{C_{score} + R(r_c + 1) - r_c}$$

Slope of the $C_{score}$ isocurves can be computed as:

$$\frac{\partial Prec}{\partial R} = \frac{1}{C_{score} + R(r_c + 1) - r_c} - \frac{(r_c + 1)R}{(C_{score} + R(r_c + 1) - r_c)^2}$$
$$= \frac{C_{score} - r_c}{(C_{score} + R(r_c + 1) - r_c)^2}$$

As described in Section III-D, these curves can have negative, positive or zero slopes depending on the value of $C_{score}$.

B. Improvements in cost for different cost ratios

Figure 7 depicts the improvement in the $C_{score}$ for different values of cost-ratios. The x-axis of the plot is $log_{10}(cost - ratio)$ and y-axis it percentage improvement in $C_{score}$ when compared to threshold selected using $F_1$ score. It can be observed that the general trend across all datasets is that the improvement is minimum near cost-ratio of 1 where $C_{score}$ behaves similar to $F_1$ score and the cost improvement increases as we move away from 1 in both the directions. The percentage increase for higher cost-ratio depends on the proportion of positive classes (class 1) in the data. This is the reason for credit card fraud there is very only minor differences in $C_{score}$ for cost-ratios greater than 1 (Figure 7b).

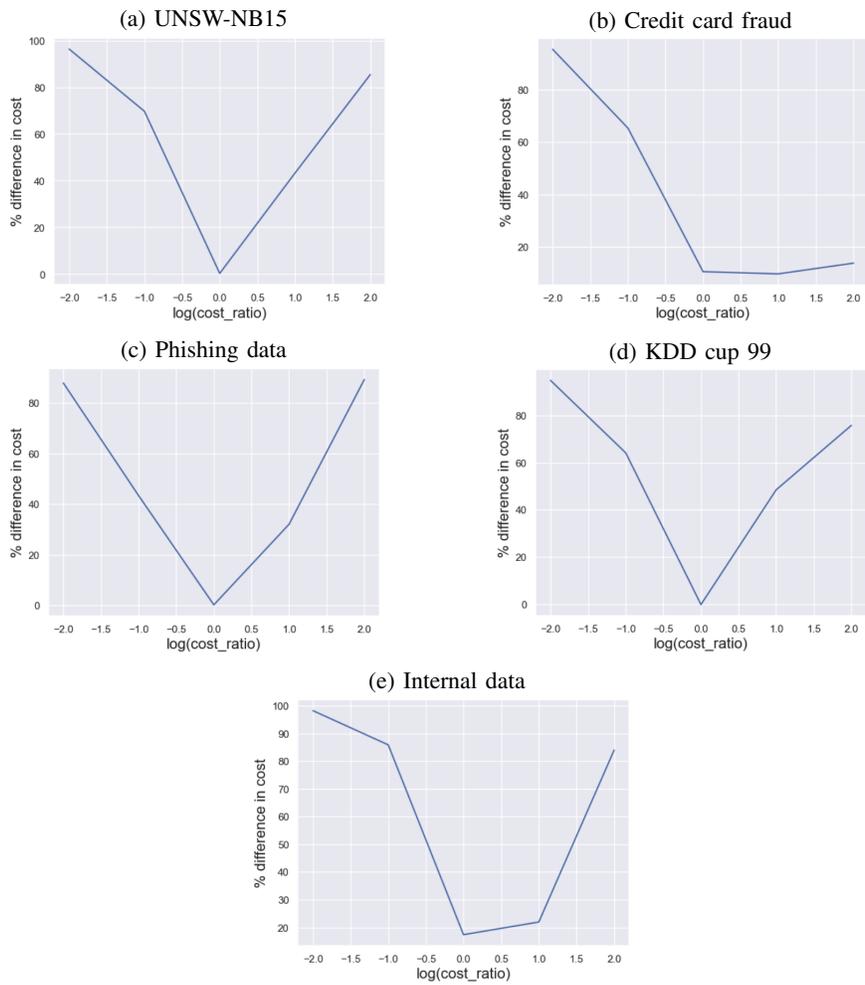

Fig. 7: Percentage improvement in cost for threshold obtained by minimizing the new cost score compared to the threshold obtained from maximizing $F_1$ score.